\documentclass[sigconf]{acmart}

\usepackage{multirow} 
\usepackage{booktabs}
\usepackage{tikz}
\usepackage{pifont}

\usepackage{colortbl}

\usepackage{cleveref}
\AtBeginDocument{%
  }

\begin{document}

\title{Multi-State Tracker: Enhancing Efficient Object Tracking via Multi-State Specialization and Interaction}

\author{Shilei Wang}
\email{shileiwang@mail.nwpu.edu.cn}
\orcid{0000-0002-0507-7256}
\affiliation{%
  \institution{School of Automation, Northwestern Polytechnical University}
  \city{Xi'an}
  \state{Shaanxi}
  \country{China}
}

\author{Gong Cheng}
\authornote{Corresponding author.}
\email{gcheng@nwpu.edu.cn}
\orcid{0000-0001-5030-0683}
\affiliation{%
  \institution{School of Automation, Northwestern Polytechnical University}
  \city{Xi'an}
  \state{Shaanxi}
  \country{China}
}

\author{Pujian Lai}
\email{laipujian@mail.nwpu.edu.cn}
\orcid{0000-0003-1357-7477}
\affiliation{%
  \institution{School of Automation, Northwestern Polytechnical University}
  \city{Xi'an}
  \state{Shaanxi}
  \country{China}
}

\author{Dong Gao}
\email{2019302284@mail.nwpu.edu.cn}
\orcid{0009-0004-5364-2194}
\affiliation{%
  \institution{School of Automation, Northwestern Polytechnical University}
  \city{Xi'an}
  \state{Shaanxi}
  \country{China}
}

\author{Junwei Han}
\email{jhan@nwpu.edu.cn}
\orcid{0000-0001-5545-7217}
\affiliation{%
  \institution{School of Automation, Northwestern Polytechnical University}
  \city{Xi'an}
  \state{Shaanxi}
  \country{China}
}
\settopmatter{printacmref=false}
\settopmatter{printacmref=false} 
\renewcommand\footnotetextcopyrightpermission[1]{} 
\pagestyle{plain} 

\begin{abstract}

Efficient trackers achieve faster runtime by reducing computational complexity and model parameters. However, this efficiency often compromises the expense of weakened feature representation capacity, thus limiting their ability to accurately capture target states using single-layer features.
To overcome this limitation, we propose Multi-State Tracker (MST), which utilizes highly lightweight state-specific enhancement (SSE) to perform specialized enhancement on multi-state features produced by multi-state generation (MSG) and aggregates them in an interactive and adaptive manner using cross-state interaction (CSI).
This design greatly enhances feature representation while incurring minimal computational overhead, leading to improved tracking robustness in complex environments.
Specifically, the MSG generates multiple state representations at multiple stages during feature extraction, while SSE refines them to highlight target-specific features. The CSI module facilitates information exchange between these states and ensures the integration of complementary features.
Notably, the introduced SSE and CSI modules adopt a highly lightweight hidden state adaptation-based state space duality (HSA-SSD) design, incurring only 0.1 GFLOPs in computation and 0.66 M in parameters.
Experimental results demonstrate that MST outperforms all previous efficient trackers across multiple datasets, significantly improving tracking accuracy and robustness. In particular, it shows excellent runtime performance, with an AO score improvement of 4.5\% over the previous SOTA efficient tracker HCAT on the GOT-10K dataset.
The code is available at https://github.com/wsumel/MST.

\end{abstract}


\maketitle

\section{Introduction}

\begin{figure}[h]
  \centering
  \includegraphics[width=1\linewidth]{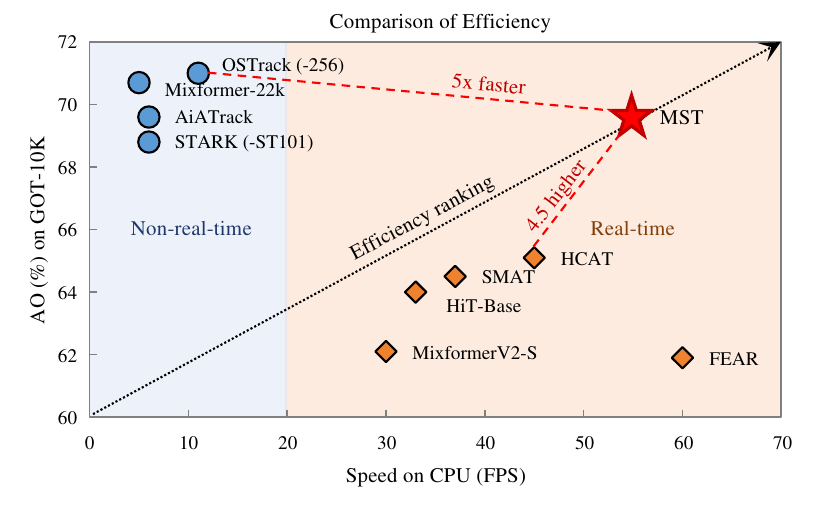}
  \caption{Comparison with the state-of-the-art trackers in
 terms of AO performance and CPU speed on GOT-10K. Following the VOT real-time standard, we define real-time performance as achieving at least 20 frames per second (FPS).}
  \label{fig:intro}
\end{figure}

Visual object tracking is a fundamental task in computer vision, with broad applications in areas such as surveillance, autonomous driving, and human-computer interaction \cite{siamrpn++,TransT,keeptrack,wsltip,lai2023ncsiam,lai2025mining,sun2024bidirectional,MCTTrack}. The goal is to accurately track an object across video frames, even under challenging conditions such as occlusions, appearance changes, and varying motion dynamics. Efficient object tracking methods are essential for real-time applications, particularly in resource-constrained environments like edge devices \cite{lighttrack, KCF,siamfc,kang2023hit,cui2024mixformerv2,wang2025cross}. These methods typically rely on lightweight models with fewer parameters to achieve fast processing speed. However, this simplification often weakens its feature extraction ability and makes it difficult to capture the complete state of the target using a single-layer representation, thus limiting its ability to handle complex tracking scenarios.

To address these limitations, we propose the Multi-State Tracker (MST), a novel tracking framework designed to enhance the robustness of efficient tracking in challenging environments. MST leverages multiple state representations to provide a more comprehensive understanding of the tracked object, enabling it to better capture variations in appearance, occlusions, and motion blur. Crucially, MST is designed to significantly improve feature representation and tracking performance while introducing minimal computational and parameter overhead, making it highly suitable for real-time and resource-constrained applications. Unlike traditional methods using a single state, MST fuses multiple states for more accurate and reliable tracking.

At the core of MST are three key modules: multi-state generation (MSG), state-specific enhancement (SSE), and cross-state interaction (CSI). The MSG module generates diverse state representations by modeling spatial relationships among patches extracted from both the template and the search region. These representations are then refined by the SSE module, which employs a global enhancement mechanism to emphasize target-specific characteristics, thereby making the tracker more sensitive to subtle variations in the object’s appearance. The CSI module facilitates information exchange between these states, integrating complementary features to further improve tracking performance.

A key advantage of MST is its efficiency. It's core modules SSE and CSI are both designed based on the hidden state adaptation-based state space duality (HSA-SSD), which features linear time complexity and minimal resource consumption. As a result, MST is capable of maintaining high tracking speeds even in real-time applications \cite{li2022odconv, lee2025efficientvim}. Despite its lightweight architecture, MST outperforms all previous efficient trackers across multiple datasets, significantly improving tracking accuracy and robustness. As shown in \Cref{fig:intro}, MST is five times faster than the traditional tracker OSTrack (-256) \cite{ostrack} and surpasses the previous best-performing efficient tracker, HCAT \cite{HCAT}, by 4.5\% in AO score on the GOT-10K \cite{got} dataset.

Our approach provides a novel solution to balancing efficiency and robustness, enabling more reliable real-time tracking in complex scenarios. The main contributions are:
\begin{itemize} \item \textbf{Multi-State Tracking Architecture:} We introduce the Multi-State Tracker (MST), which leverages multiple state representations to enhance tracking accuracy and robustness, enabling better handling of variations in appearance, occlusions, and motion blur. \item \textbf{Key Technical Innovations:} We propose three key modules Multi-State Generation (MSG), State-Specific Enhancement (SSE), and Cross-State Interaction (CSI), with the latter two built upon the Hidden State Adaptation-based State Space Duality (HSA-SSD). Together, these modules enable the generation of diverse state-aware features, the refinement of individual state representations, and effective information exchange across states. Importantly, they enhance tracking performance while introducing only minimal computational overhead. \item \textbf{State-of-the-Art Performance:} Leveraging its innovative architectural design and key advancements, MST not only maintains exceptional processing speed but also delivers strong tracking performance, as demonstrated by extensive experimental evaluations across several benchmark datasets. \end{itemize}

\section{Related Work}

\textbf{Efficient Object Tracking.} Efficient object tracking has become increasingly important for real-world applications, particularly on edge devices. Early methods, such as ECO \cite{danelljan2017eco} and ATOM \cite{atom}, demonstrated real-time capabilities, but their tracking accuracy often fell short. Subsequent efforts have sought to better balance speed and performance. For example, LightTrack \cite{lighttrack}, FEAR \cite{borsuk2022fear}, and HCAT \cite{HCAT} have explored lightweight network architectures and efficient design strategies to reduce computational complexity while still delivering competitive performance. With the advent of one-stream architectures, one-stream efficient tracking frameworks such as MixFormerV2 \cite{cui2024mixformerv2} and HiT \cite{kang2023hit} integrate feature extraction and interaction into a unified process, achieving impressive accuracy through techniques like distillation, pruning, and innovative architectural design. However, these methods all rely on a single-layer feature representation to capture the target's state. In contrast, our proposed Multi-State Tracker (MST) leverages multiple state representations and interactive aggregation to robustly capture diverse target variations, thereby significantly enhancing tracking robustness while maintaining extremely high computational efficiency.

\noindent\textbf{State Space Model.} State Space Models (SSMs) \cite{gu2023mamba} have recently gained considerable attention as an efficient and scalable solution for sequence modeling, offering linear time complexity while capturing long-range dependencies. These models, originating from the Kalman filter, have been widely adopted in various domains, particularly in natural language processing (NLP), where models like S4, DSS, and Mamba have demonstrated their ability to handle structured state transitions and allow for efficient parallel computation. The ability to efficiently model temporal dependencies while maintaining low computational overhead has made SSMs a promising candidate for various sequence-based tasks.

\begin{figure*}[h]
  \centering
  \includegraphics[width=1\linewidth]{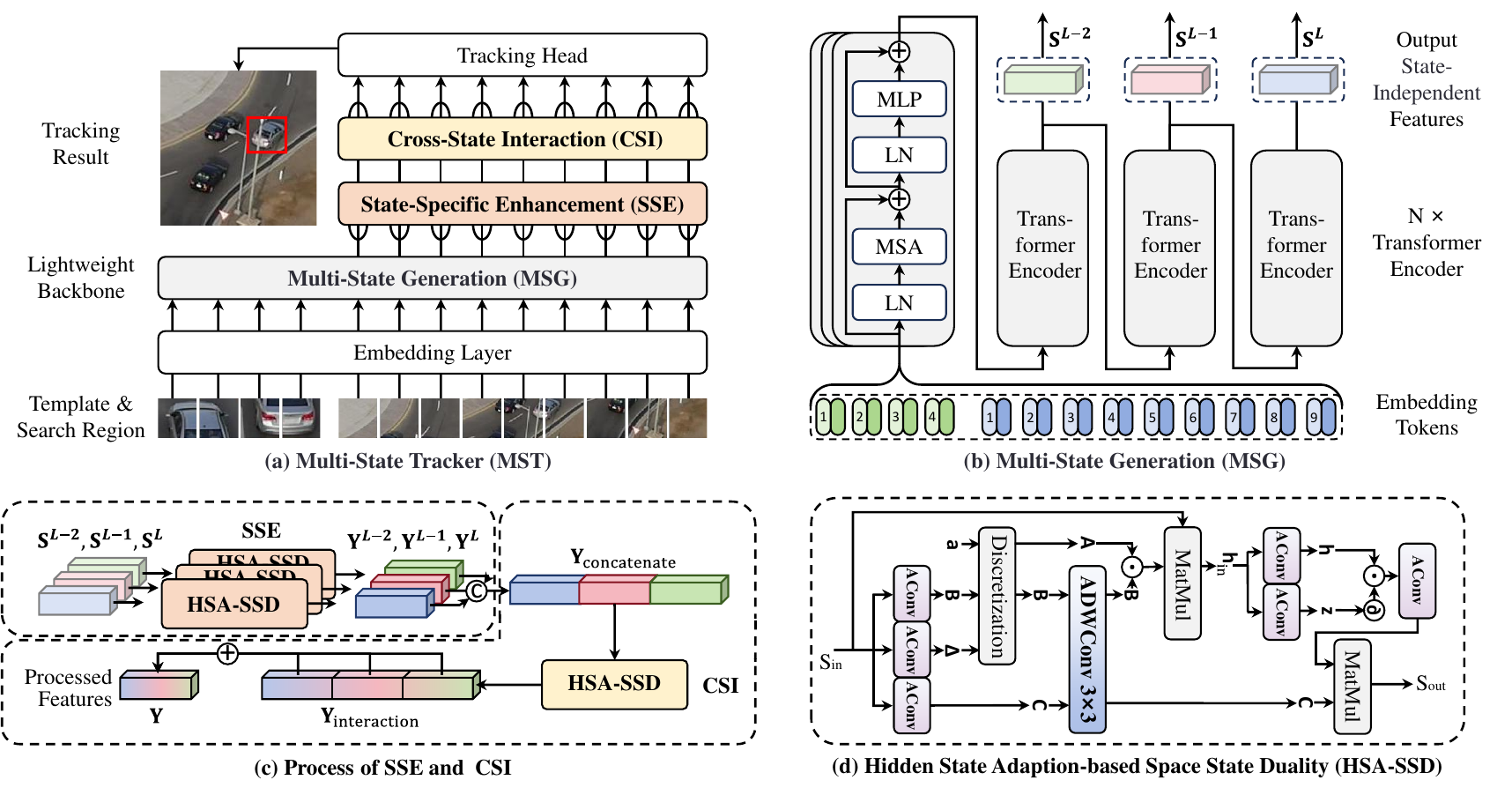}
  \caption{(a) Overview of the proposed Multi-State Tracker (MST). (b) The Multi-State Generation (MSG) outputs three potential states of the tracked object in the final three layers. (c) The Process of  state-specific enhancement (SSE) and  cross-state interaction (CSI). (d) The architecture of hidden state adaption-based state space duality (HSA-SSD).}
  \label{fig:method}
\end{figure*}

In the field of computer vision, the use of SSMs has begun to gain traction as well. Vision-based adaptations such as Vim \cite{lee2025efficientvim}, VMamba \cite{liu2024vmamba}, and MambaVision \cite{hatamizadeh2024mambavision} have explored the application of SSMs as a backbone for visual tasks. For instance, Vim employs bidirectional state space models enhanced with position embeddings to facilitate better visual representation learning, while VMamba introduces a 2D Selective Scan (SS2D) mechanism, combining the sequential nature of SSMs with the spatial complexity of vision data. In the object tracking domain, previous work MCITrack~\cite{mcitrack} employed Mamba SSMs within a ViT-style backbone to model long-term sequence dependencies, enabling robust tracking over extended frames.

\noindent\textbf{State Space Duality (SSD).} 
State Space Duality (SSD) was first introduced by researchers in Mamba-2\cite{dao2024mamba2}, featuring a core layer that enhances Mamba's selective state space model (SSM). This improvement not only further accelerates processing speed but also maintains competitive performance with SSM in long-sequence modeling.
Traditional SSD formulations, however, often rely on causal structures inherited from their original use in sequence modeling, which can be suboptimal for spatially dense vision inputs.
To overcome this, Non-Causal SSD (NC-SSD)  \cite{ncssd} was proposed by removing the causality constraint, thereby enabling bidirectional information flow and improving global context modeling in vision scenarios. NC-SSD computes global hidden states through weighted combinations of input tokens and uses them to generate the output via linear projections. Nonetheless, the computational cost remains high due to the per-token application of gating and output projections, which scale quadratically with the channel dimension.
To further enhance efficiency, Hidden State Mixer-based SSD (HSM-SSD)  \cite{lee2025efficientvim} was introduced as a refinement of NC-SSD \cite{ncssd}. It restructures the computation around a compact set of global hidden states, significantly reducing cost by performing the gating and output projection directly in the reduced latent space. This design lowers the complexity from $\mathcal{O}(LD^2 + LND)$ to $\mathcal{O}(ND^2 + LND)$, where $N \ll L$, without compromising global modeling capacity. Additionally, a single-head version of HSM-SSD with state-wise importance weights has been shown to reduce memory overhead while maintaining performance comparable to multi-head designs.

Despite these advances, existing SSD-based vision models have mainly concentrated on image classification task, leaving visual object tracking relatively underexplored. To bridge this gap, we propose an improved variant of HSM-SSD, termed hidden state adaptation-based state space duality (HSA-SSD), which is tailored to better accommodate diverse input state features encountered in tracking scenarios. Based on this, we further develop a multi-state representation fusion strategy which combines feature specialization enhancement and interactive aggregation mechanism. These enhancements significantly improve the representation capability of lightweight trackers while maintaining computational efficiency. As a result, the proposed framework achieves strong robustness under complex tracking conditions, making MST a powerful solution that effectively connects high-performance visual tracking with real-time deployment.

\section{Method}

This section begins with an overview of the Multi-State Tracker (MST) in \Cref{ss3.1}, followed by a discussion of the multi-state generation in \Cref{ss3.2}. \Cref{ss3.3} presents the state-specific enhancement and cross-state interaction modules, which improve state features specificity and enable information exchange. Finally, \Cref{ss3.4} outlines the implementation of the tracking head.

\subsection{Overview}
\label{ss3.1}

\Cref{fig:method}a illustrates the overall architecture of the proposed Multi-State Tracker (MST). The process begins by embedding the template and search region to generate tokenized representations. These tokens are then concatenated and fed into the multi-state generation (MSG) module. The MSG module is responsible for modeling the relationships between tokens and performing multi-level features extraction, effectively capturing different states of the tracked target.
Next, these features are fed into the state-specific enhancement (SSE) module. In this stage, we adopt a global modeling ability enhancement mechanism driven by hidden state adaptation-based spatial state duality (HSA-SSD) to refine each features. This process enhances the ability of features to express specific states, allowing the tracker to better represent subtle changes in the target.
Following the enhancement, the refined features are passed to the cross-state interaction (CSI) module. Here, information is exchanged between multiple features, enabling the mutual reinforcement and integration of complementary feature across different representations. Finally, the outputs from the CSI module are aggregated into a single unified feature, which is then forwarded to the tracking head for the final prediction. This multi-state representation design enables MST to ensure robust and accurate tracking even in complex scenes while keeping the model lightweight and efficient.

\subsection{Multi-State Generation}
\label{ss3.2}

The multi-state generation (MSG) module, shown in \Cref{fig:method}b, is designed to capture diverse target representations by processing both the template and search region in a unified manner. To achieve this, the template $\textbf{Z} \in \mathbb{R}^{B \times M \times C}$ and search region $\textbf{X} \in \mathbb{R}^{B \times N \times C}$ are first partitioned into multiple patches. Each patch is then projected into a latent space through a linear transformation:  

\begin{equation}  
\textbf{p}_i = W_p \textbf{v}_i + b_p, \quad \text{where} \quad \textbf{v}_i \in \{\textbf{x}_i, \textbf{z}_i\},
\end{equation}  
where $W_p \in \mathbb{R}^{d' \times d}$ is the projection matrix, $b_p \in \mathbb{R}^{d'}$ is the bias term, and $d'$ represents the dimension of the latent space. Here, $\{\textbf{x}_i, \textbf{z}_i\}$ denotes a patch obtained from the partitioned template $\textbf{Z} \in \mathbb{R}^{B \times M \times C}$ or search region $\textbf{X} \in \mathbb{R}^{B \times N \times C}$.  

Next, the MSG module takes the concatenated tokens obtained from the embedded template $\textbf{Z}$ and search region $\textbf{X}$ as input. Let $\textbf{P}_{[\text{Z;X}]} = \{\textbf{p}_{z1}, \textbf{p}_{z2}, \dots, \textbf{p}_{zM}; \textbf{p}_{x1}, \textbf{p}_{x2}, \dots, \textbf{p}_{xN}\}$ denote the set of these tokens, where each token $\textbf{p}_{zi},\ \textbf{p}_{xi} \in \mathbb{R}^{d}$ is a $d$-dimensional feature vector.


Within this latent space, the concatenated tokens $\textbf{P}_{[\text{Z;X}]}$ are sequentially processed through multiple hierarchical blocks to model spatial contextual relationships. Each block consists of two layer normalization (LN), a multi-head self-attention (MHSA) layer, and a multilayer perceptron (MLP). By stacking multiple such blocks, the MSG module effectively captures long-range dependencies among patches and refines feature representations.  

Specifically, the MSG module uses multi-level attention blocks to extract features and model relationships. The attention mechanism in the block is defined as:

\begin{equation}
\text{Attention}(\textbf{Q}_{\text{Z;X}}, \textbf{K}_{\text{Z;X}}, \textbf{V}_{\text{Z;X}}) = \text{softmax}\left(\frac{\textbf{Q}_{\text{Z;X}}\textbf{K}_{\text{Z;X}}^\top}{\sqrt{d_k}}\right)\textbf{V}_{\text{Z;X}},
\end{equation}  
where $\textbf{Q}_{\text{Z;X}}$, $\textbf{K}_{\text{Z;X}}$ and $\textbf{V}_{\text{Z;X}}$ are query, key, and value matrices replicated by projection tokens $\textbf{P}_{[\text{Z;X}]}$, and $d_k$ is the dimensionality of each attention head. This mechanism enables the search region features to iteratively integrate different representations of the target’s state under the guidance of template cues.  

To capture diverse target representations, we extract state features from the last three layers of the MSG module. Let $\textbf{S}^{(L-2)}$, $\textbf{S}^{(L-1)}$, and $\textbf{S}^{(L)}$ denote the feature representations from layers $L-2$, $L-1$, and $L$, respectively. These multi-scale representations encode different aspects of the target’s appearance and are subsequently forwarded for further processing.  
\subsection{State Specialization and Interaction}  
\label{ss3.3}  

The process of state-specific enhancement and interaction is illustrated in \Cref{fig:method}c. Initially, the multi-state features \( \textbf{S}^{L-2}, \textbf{S}^{L-1}, \textbf{S}^L \), which are independently generated by the multi-state generation (MSG), are fed into the state-specific enhancement (SSE) module in parallel. Each state features are processed individually to enhance its representation.

\begin{equation}
 \textbf{Y}^{L-2}, \textbf{Y}^{L-1}, \textbf{Y}^L = \text{SSE}(\textbf{S}^{L-2}), \text{SSE}(\textbf{S}^{L-1}), \text{SSE}(\textbf{S}^L).
\end{equation}

The SSE module consists of multiple independent state space model blocks, each of which performs bidirectional relationship modeling on the input state features, thereby enhancing their distinctiveness and ensuring that each state captures unique information.

After refinement, the enhanced state features \( \textbf{Y}^{L-2}, \textbf{Y}^{L-1}, \textbf{Y}^L \) are concatenated to form a unified state representation \( \textbf{Y}_{\text{concatenate}} \), which integrates the information from all states into a single feature vector:

\begin{equation}
    \textbf{Y}_{\text{concatenate}} = \text{Concatenate}(\textbf{Y}^{L-2}, \textbf{Y}^{L-1}, \textbf{Y}^L).
\end{equation}

This joint representation is then passed through the cross-state interaction (CSI) module, which models bidirectional relationships across the different states. The CSI module ensures that each state attends to the others, allowing the model to capture global inter-state dependencies.

\begin{equation}
    \textbf{Y}_{\text{interaction}} = \text{CSI}(\textbf{Y}_{\text{concatenate}} ).
\end{equation}

Finally, the interacted state representation \( \textbf{Y}_{\text{interaction}} \) is split back into its original components based on the concatenation order. These components are then element-wise summed to produce the final enhanced state representation \( \textbf{Y} \), which captures the integrated features from all states:

\begin{equation}
    \textbf{Y} = \textbf{Y}_{\text{interaction}}^{L-2} + \textbf{Y}_{\text{interaction}}^{L-1} + \textbf{Y}_{\text{interaction}}^L.
\end{equation}

\noindent\textbf{State-Specific Enhancement (SSE):}  
The SSE module refines state representations by explicitly modeling directional dependencies within each state. 
The process in SSE is shown in \Cref{fig:method}d.
Given the input sequence $
\textbf{S}^i = \{\textbf{S}^{(L-2)}, \textbf{S}^{(L-1)}, \textbf{S}^{(L)}\},
$
The sequence is then input into an adaption convolution1 $\times$ 1, which dynamically adjusts the weights according to the specific feature patterns of different states, making the model more adaptable to different state representations. The mathematical operation is as follows:

\begin{equation}
\textbf{B}, \textbf{C}, \Delta = \text{AConv1 $\times$ 1}(\textbf{S}^i),    
\end{equation}
where the adaption convolution (AConv)1 $\times$ 1 is adaptively adjusted based on the attention weights calculated based on the input state. The process can be expressed mathematically as:

\begin{equation}
\tilde{W}(\textbf{S}^i) = \sum_{k=1}^K \pi_k(\textbf{S}^i) \tilde{W}_k,
\end{equation}
where \( \tilde{W}_k \) are the1 $\times$ 1 convolution kernels, and \( \pi_k(\textbf{S}^i) \) represents the attention weight for the \( k \)-th kernel computed from the input state \( \textbf{S}^i \). The attention weights \( \pi_k(\textbf{S}^i) \) are determined by a lightweight network, typically using global pooling and an MLP layer.

\begin{equation}
\textbf{B}, \textbf{C}, \Delta = \tilde{W}(\textbf{S}^i) \ast \textbf{S}^i + \tilde{b}(\textbf{S}^i),
\end{equation}
where \( \ast \) denotes the convolution operation, and \( \tilde{b}(x) \) is a dynamic bias term. This step allows the model to capture inter-channel dependencies and adapt to different feature patterns in the input states. The attention mechanism assigns different importance to the kernels, enabling the model to focus on the most relevant features of the input. The key benefit of using AConv1 $\times$ 1 is its ability to process diverse state characteristics in an adaptive manner, thereby improving the power of SSD in handling various state representations.

After this, the matrices \( \textbf{B} \) and \( \textbf{C} \) are processed using ADWConv 3 $\times$ 3. This step allows the model to learn local spatial dependencies, making it more capable of understanding the intricate spatial relationships in the input data. The operation is:

\begin{equation}
\textbf{B}, \textbf{C} = \text{ADWConv 3 $\times$ 3}(\textbf{B}, \textbf{C}),    
\end{equation}
where adaptation depthwise convolution (ADWConv) $3 \times 3$ applies $3 \times 3$ convolutions across the spatial dimensions, capturing local spatial dependencies within the state feature maps.

Next, the matrices \( \textbf{a} \) and \( \textbf{B} \) are discretized using the matrix \( \Delta \), which results in the hidden state \( \mathbf{h}_{\text{in}} \) being computed by multiplying the discretized matrices \( \textbf{A} \) and \( \textbf{B} \) with the input \( \mathbf{\textbf{S}}_{\text{in}}^i \):

\begin{equation}
    \mathbf{h}_{\text{in}}^i = \textbf{A} \odot \textbf{B}^\top \mathbf{\textbf{S}}_{\text{in}}^i,
\end{equation}

where  \( \mathbf{\textbf{S}}_{\text{in}}^i \) is the input sequence at the current time step. 

The hidden states are then passed through a AConv1 $\times$ 1 transformation and gated with a nonlinear activation function \( \sigma \). This can be expressed as:

\begin{equation}
\mathbf{h}_{\text{out}}^i = \text{AConv1 $\times$ 1}(\sigma \odot \mathbf{h}_{\text{in}}^i).
\end{equation}

Finally, the output \( \mathbf{Y}^i \) is computed by projecting the hidden states and performing a Hadamard product with the matrix \( \textbf{C} \), yielding the final enhanced state representation:

\begin{equation}
\mathbf{Y}^i = \textbf{C} \cdot \mathbf{h}_{\text{out}}^i.
\end{equation}

The use of AConv1 $\times$ 1 in both the projection and gating steps ensures that the model can adaptively capture both inter-channel and spatial dependencies, making it well-suited for handling varying state representations.

\noindent\textbf{Cross-State Interaction (CSI):}  
The CSI module integrates the features from multiple states to capture global inter-state relationships. The concatenated state representations from the SSE module, \( \textbf{Y}_{\text{concatenate}} = \text{Concatenate}(\textbf{Y}^{L-2}, \textbf{Y}^{L-1}, \textbf{Y}^{L}) \), are processed by the same steps as the SSE module, involving AConv1 $\times$ 1 for linear projections, ADWConv 3 $\times$ 3 for capturing local spatial dependencies, and discretization. After this, the features are passed through a linear transformation and nonlinear activation functions to capture the local dependencies within the concatenated states. The global state dependencies are then modeled through the Hadamard product operation, and the interactive state features $\textbf{Y}_{\text{interaction}} $ are output.
Then the parts are split and added together to get the final state features $\textbf{Y}$:

\begin{equation}
    \textbf{Y} = \textbf{Y}_{\text{interaction}}^{L-2} + \textbf{Y}_{\text{interaction}}^{L-1} + \textbf{Y}_{\text{interaction}}^L.
\end{equation} 

\textbf{Y} is used as the feature input to the tracking head after specialization and interaction enhancement for tracking.

\subsection{Tracking Head}
\label{ss3.4}
We employ a center head to estimate the target’s centroid and scale. The center head consists of three parallel branches, each composed of multiple Conv-BN-ReLU layers. These branches generate three essential outputs: a classification score map \( \textbf{P} \in [0,1]^{H_P \times W_P} \), which encodes the likelihood of the target’s presence at each spatial location; a bounding box size map \( \textbf{B} \in [0,1]^{2 \times H_P \times W_P} \), which predicts the normalized width and height of the target; and an offset map \( \textbf{O} \in [0,1)^{2 \times H_P \times W_P} \), designed to refine localization by mitigating discretization errors.

The target position is determined by selecting the location with the highest classification score:

\begin{equation}
(x_d, y_d) = \arg\max_{(x,y)} \textbf{P}_{x,y}.
\end{equation}

The final bounding box is computed by incorporating the predicted offsets and bounding box size:

\begin{equation}
(x, y, w, h) = (x_d + \textbf{O}^{(0)}_{x_d, y_d}, y_d + \textbf{O}^{(1)}_{x_d, y_d}, \textbf{B}^{(0)}_{x_d, y_d}, \textbf{B}^{(1)}_{x_d, y_d}).
\end{equation}

For training, we optimize both classification and regression objectives. The classification branch employs a weighted focal loss to mitigate class imbalance, while the regression branch utilizes a combination of \( \ell_1 \) loss and generalized IoU loss [40] to enhance localization precision. The overall loss function is formulated as:

\begin{equation}
\mathcal{L}_{\text{track}} = \mathcal{L}_{\text{cls}} + \lambda_{\text{iou}} \mathcal{L}_{\text{iou}} + \lambda_{L1} \mathcal{L}_{L1},
\end{equation}
where the regularization parameters are set to \( \lambda_{\text{iou}} = 2 \) and \( \lambda_{L1} = 5 \) to balance the contributions of different loss terms, following prior works

\section{Experiment}

\begin{table*}

  \caption{Compare MST with the state-of-the-art methods using three large-scale benchmark datasets, including GOT-10k~\cite{got}, TrackingNet~\cite{trackingnet}, LaSOT~\cite{lasot}. The best results are highlighted in \textcolor{red}{red} font.}
\resizebox{1\textwidth}{!}{
  \centering
  
  \begin{tabular}{l|l|l|ccc ccc ccc cc}
    \toprule[1.5pt]
    \textcolor{white}{aaa}&\multirow{2}{*}{Tracker} 
    & \multirow{2}{*}{Publication}
    
    & \multicolumn{3}{c}{GOT-10k \cite{got} }  & 

    \multicolumn{3}{c}{TrackingNet \cite{trackingnet}}  & \multicolumn{3}{c}{LaSOT \cite{lasot}}  & \multicolumn{2}{c}{ FPS }  \\
    
      & & & AO & $\text{SR}_{0.5}$ & SR$_{0.75}$ & AUC &  P$_\text{{Norm}}$ & P & AUC &  P$_{\text{Norm}}$ & P  & GPU & CPU \\
    \midrule[0.75pt]
\multirow{11}{*}{\rotatebox{90}{Lightweight tracking}} &

\textbf{MST} & Ours & \textcolor{red}{69.6} & \textcolor{red}{79.8} & \textcolor{red}{64.1} & \textcolor{red}{81.0} & \textcolor{red}{86.1}  &  \textcolor{red}{78.7} & \textcolor{red}{65.8} & \textcolor{red}{75.2} & \textcolor{red}{70.1} & 200 & 55 \\


\cmidrule(lr){2-14}

& AsymTrack-B ~\cite{AsymTrack} & AAAI 2025 & 67.7 & 76.6 & 61.4 &80.0 & 84.5 & 77.4 & 64.7 &73.0 & 67.8 & 197& 38 \\

& SMAT \cite{Gopal_2024_WACV} & WACV2024 & 64.5 &74.7 &57.8 & 78.6 &84.2 &75.6 &61.7 &71.1 & 64.6 & 158 & 37 \\

&HiT-Base \cite{kang2023hit} & ICCV2023 & 64.0 & 72.1 & 58.1    &  80.0 & 84.4 & 77.3     &  64.6 & 73.3 & 68.1 & 175  & 33  \\
&HiT-Small \cite{kang2023hit} & ICCV2023 & 62.6 & 71.2 & 54.4   &    77.7 & 81.9 & 73.1       &  60.5 & 68.3 & 61.5 & 192  & 72  \\
&HiT-Tiny \cite{kang2023hit} & ICCV2023 & 52.6 & 59.3 & 42.7   &  74.6 & 78.1 & 68.8    &  54.8 & 60.5 & 52.9   &  204 & \textcolor{red}{76}  \\
&MixformerV2-S \cite{cui2024mixformerv2} &NeurIPS2023 & 62.1 & - & - & 75.8 & 81.1 & 70.4 & 60.6 & 69.9 
 & 60.4 & \textcolor{red}{325} & 30 \\  

&E.T.Track \cite{ETTrack} & WACV2023 & -& -&-     &   75.0 & 80.3 & 70.6    &  59.1 & - & -  & 40  & 47  \\

&FEAR \cite{borsuk2022fear} & ECCV2022 & 61.9 & 72.2 & -    &  - &  - &   -    &   53.5 & - & 54.5   & 105  & 60  \\

&HCAT \cite{HCAT} & ECCVW2022 & 65.1 & 76.5 & 56.7    &   76.6 & 82.6 & 72.9      & 59.3 & 68.7 & 61.0  &  195 &  45 \\

&LightTrack \cite{lighttrack} & CVPR2021 &61.1 & 71.0 & -    &   72.5 & 77.8 & 69.5      &  53.8 & - & 53.7 &  128 & 41  \\



    \cmidrule(lr){1-14}
\multirow{9}{*}{\rotatebox{90}{Heavyweight tracking}} &

MCITrack-B224   ~\cite{MCTTrack} & AAAI2025 &77.9 & 88.2 & 76.8 & 86.3 & 90.9 & 86.1 & 75.3 & 85.6 & 83.3  &35 & 5 \\

&SAMURAI-L  ~\cite{yang2024samurai} & Arxiv2024 & 81.7 & 92.2 & 76.9 & 85.3 & - & - & 74.2 & 82.7 & 80.2  &- & - \\

&ODTrack   ~\cite{zheng2024odtrack} & AAAI2024 & 77.0 & 87.9 & 75.1 & 85.1 & 90.1 & 84.9 & 73.2 & 83.2 & 80.6  &32 & 5 \\

&ARTrack-256  ~\cite{artrack} & CVPR2024& 73.5 & 82.2 & 70.9 & 84.2 & 88.7 & 83.5 & 70.4 & 79.5 & 76.6  &37 &6\\

&OSTrack (-256)~\cite{ostrack}        & CVPR2022 & 71.0 & 80.4 & 68.2 &  83.1 & 87.8 & 82.0 &  69.1 & 78.7 & 75.2  & 105  & 11   \\

&AiATrack~\cite{gao2022aiatrack}           & ECCV2022 & 69.6   & 63.2 & 80.0 &  82.7 & 87.8 & 80.4   & 69.0  & 79.4   & 73.8  &   38 &  6  \\

&Mixformer-22k~\cite{mixformer}      & CVPR2022 & 70.7   & 80.0 & 67.8 &   83.1 & 88.1 & 81.6       & 69.2  & 78.7   & 74.7  &   25 &  5  \\
&ToMP (-101)~\cite{tomp}        & CVPR2022 & -      & -    & -    & 81.5 & 86.4      & 78.9      & 68.5  & 79.2   & 73.5  &  25  & 5   \\

&Stark (-ST101)  \cite{stark}      & ICCV2021 & 68.8   & 78.1 & 64.1 &   82.0  & 86.9       &     -   & 67.1  & 76.9   & -      & 32   & 5  \\




    \bottomrule[1.5pt]
  \end{tabular}}

  \label{tab01}

\end{table*}

\subsection{Implementation Details}

\noindent\textbf{Model.} 
We propose the Multi-State Tracker (MST), which is built upon a ViT-Tiny backbone~\cite{attention,ViT} and initialized with distilled pre-trained weights from MAE~\cite{he2022mae,wang2023kdmae}. Our framework is designed to operate on inputs with a template size of 128$~\times~$128 and a search region size of 256$~\times~$256, enabling efficient and robust target tracking across challenging scenarios.

\noindent\textbf{Training.} 
Our MST is developed and trained using PyTorch 1.8.1 with Python 3.9.19, leveraging four NVIDIA RTX 2080Ti GPUs. The training dataset is assembled from several established benchmarks, including LaSOT~\cite{lasot}, TrackingNet~\cite{trackingnet}, GOT-10K~\cite{got}, and COCO2017~\cite{coco2017}. In particular, the GOT-10K protocol is strictly followed by utilizing only its designated training split.
The network optimization is carried out using the AdamW optimizer over 300 epochs, with each epoch comprising 60,000 image pairs. To curb overfitting, the training on the GOT-10K benchmark is confined to 100 epochs.

\begin{table}
\caption{Comparison with SOTA lightweight trackers in terms of FLOPs and Params. The best results are shown in \textcolor{red}{red} font.}
  \centering
  \resizebox{1\linewidth}{!}{
  \begin{tabular}{c|c|ccc}
    \toprule
    
    Method & Publication &FLOPs(G) & Params(M) & UAV123 \\
    \midrule
    \textbf{MST} & Ours & 2.28  &  7.80 & \textcolor{red}{68.4} \\
    AVTrack~\cite{AVTrack} & ICML2024 & 0.97-2.4 & 3.5-7.9 & 66.8 \\
    HiT-Base~\cite{kang2023hit} & ICCV2023 &4.34 &42.14 & 65.6 \\
    MixformerV2-S~\cite{cui2024mixformerv2} & NeurIPS2023 &4.4 &16.04 & 65.1 \\
    \bottomrule
  \end{tabular}}
  \label{tab:param}
\end{table}

\noindent\textbf{Inference.} 
During the inference stage, positional priors are incorporated into tracking by performing element-wise multiplication between the classification response map and a Hanning window of the same dimensions. The candidate region with the highest adjusted score is then selected as the final tracking position, determining the ultimate bounding box.

As shown in \Cref{tab:param}, we compare the computational cost (FLOPs) and parameters (Params) of our MST with state-of-the-art lightweight trackers. MST requires only half the FLOPs of the classical MixFormerV2-S and HiT-Base while achieving AUC the gains of 3.3\% and 2.8\% on the UAV dataset. Compared to the latest AVTrack, MST maintains a similar FLOPs and parameter count but outperforms it by 1.6\% in AUC on UAV123 dataset, demonstrating superior efficiency and tracking performance.

\subsection{Comparison with State-of-the-arts}

We conduct a comprehensive evaluation of our proposed MST against state-of-the-art methods across seven benchmark datasets: GOT-10K, TrackingNet, LaSOT, TNL2K, UAV123, NFS, and LaSOT$_{\text{ext}}$. Notably, we categorize the trackers into lightweight and heavyweight groups.

\noindent\textbf{GOT-10K.} The GOT-10K dataset ~\cite{got} is a large-scale benchmark for object tracking with different training and test splits. To ensure a rigorous and fair evaluation of tracking performance, the tracker is required to be trained only using data from the training set.
As shown in ~\Cref{tab01}, MST achieves outstanding results on GOT-10K, surpassing the previous state-of-the-art performance by 4.5\%, 3.3\% and 6.0\% in AO, SR$_{0.5}$ and SR$_{0.75}$, respectively, reaching the scores of 69.6\%, 79.8\% and 64.1\%.
This result fully demonstrates the effectiveness of MST in enhancing target perception and tracking robustness. By utilizing multiple state features to represent the target, MST improves the positioning accuracy and adaptability to complex scenes.

\noindent\textbf{TrackingNet.} We further evaluate the trackers on the 511 test sequences from the TrackingNet dataset, which contains a total of 30643 sequences, with 30132 used for training and 511 reserved for testing ~\cite{trackingnet}. 
The results are presented in ~\Cref{tab01}.
Compared to the previous best lightweight tracker, HiT-Base, MST achieves notable improvements of 1.0\%, 1.7\%, and 1.4\% in AUC, P$_{\text{Norm}}$, and P, respectively, reaching new state-of-the-art scores of 81.0\%, 86.1\%, and 78.7\%. This further validates MST's exceptional tracking capabilities, highlighting its superior accuracy, robustness, and adaptability across diverse and challenging scenarios.

\noindent\textbf{LaSOT.} The LaSOT dataset serves as a benchmark for long-term tracking, featuring 280 test sequences that cover 14 diverse object categories ~\cite{lasot}. With an average of 2500 frames per sequence, it poses significant challenges for tracking algorithms. Given its extended duration and varied complexities, achieving high performance requires exceptional robustness.
As shown in ~\Cref{tab01}, MST also achieves the best performance among lightweight trackers on the LaSOT dataset, with AUC, P$_{\text{Norm}}$, and P scores of 65.8\%, 75.2\%, and 70.1\%, respectively. It surpasses the previous best tracker, HiT-Base, by 1.2\%, 1.9\%, and 2.0\%, while showing even greater improvements over SMAT, with gains of 4.1\%, 4.1\%, and 5.5\%.

\begin{figure}[t]
  \centering
   \includegraphics[width=1\linewidth]{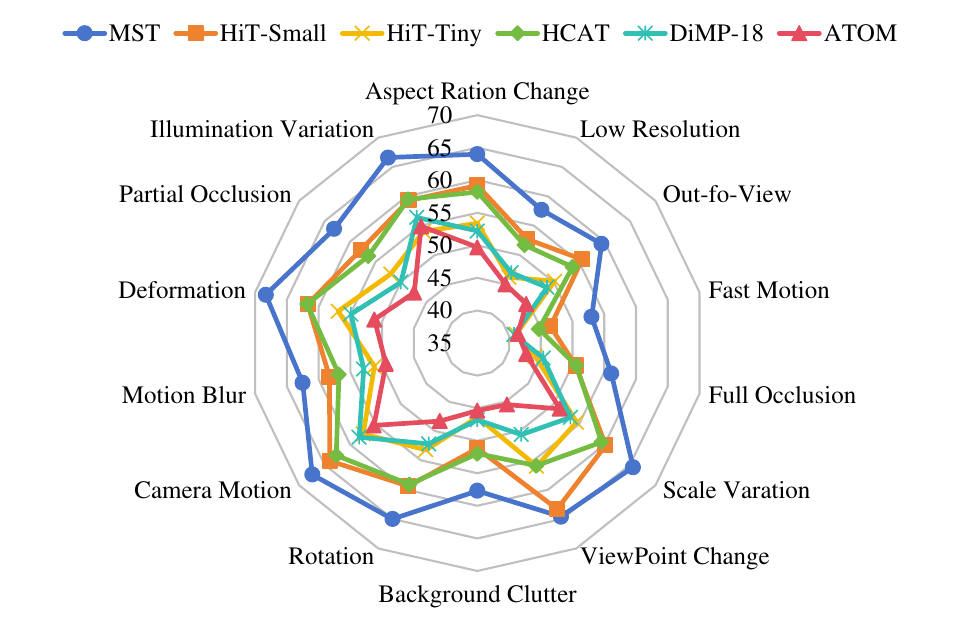}

   \caption{AUC of different attributes on LaSOT.}
   \label{fig:att}
\end{figure}

We compare our tracker against computationally comparable methods on various attributes of the LaSOT dataset. As shown in \Cref{fig:att}, the proposedd MST, achieves a clear and consistent advantage over competing trackers, outperforming all others across all 14 attributes. This strong performance demonstrates the effectiveness of the proposed SSE and CSI modules, built upon the hidden state adaptation-based state space duality (HSA-SSD), in enhancing the tracker’s robustness and accuracy in complex scenarios.


\noindent\textbf{TNL2K, UAV123, NFS and LaSOT$_{\text{ext}}$.} 
To demonstrate the robustness of our MST across a diverse range of scenarios, we evaluated them on four additional benchmarks:
1) TNL2K: A multimodal dataset featuring 700 video sequences, it integrates natural language annotations and includes challenging scenarios with pedestrian appearances undergoing cloth or face alterations ~\cite{tnl2k}.
2) UAV123: An aerial dataset consisting of long-term video sequences captured in complex environments, highlighting challenges associated with UAV-based tracking ~\cite{UAV}.
3) NFS: A high-speed dataset with videos recorded at 240 frames per second (FPS), specifically designed to test tracking performance under rapid motion ~\cite{NFS}.
4) LaSOT$_{\text{ext}}$: An extended version of LaSOT, comprising 150 long-term video sequences, providing additional challenges for long-duration tracking tasks ~\cite{fan2021lasotext}.
As shown in \Cref{tunl}, MST achieves the highest AUC scores across all four datasets, reaching 53.3\%, 68.4\%, 65.4\%, and 45.1\%. Compared to the previous best method, it demonstrates the improvements of 6.1\%, 1.6\%, 1.8\%, and 1.0\%, respectively.
These results further emphasize MST’s strong generalization ability, showcasing its effectiveness in maintaining stable and precise tracking performance across a wide range of complex conditions.

\begin{table}
\caption{\textcolor{black}{Compare different models on AUC of TNL2K ~\cite{tnl2k}, UAV123 ~\cite{UAV}, NFS ~\cite{NFS} and LaSOT$_{\text{ext}}$ ~\cite{fan2021lasotext}. The best results are highlighted by \textcolor{red}{red} font.}}
\centering
\resizebox{1\linewidth}{!}{
\begin{tabular}{l|c|cccc}
 \toprule[1.5pt]
Method & Publication &\begin{tabular}[c]{@{}c@{}} TNL2K \\  \end{tabular} & \begin{tabular}[c]{@{}c@{}}  UAV123 \\  \end{tabular}  &  \begin{tabular}[c]{@{}c@{}} NFS \\ \end{tabular} & LaSOT$_{\text{ext}}$ \\
 \midrule[0.75pt]


 \textbf{MST} & our & \textcolor{red}{53.3} & \textcolor{red}{68.4} & \textcolor{red}{65.4} & \textcolor{red}{45.1} \\
  \midrule[0.75pt]

 AVTrack-DeiT \cite{AVTrack} & ICML2024 & - & 66.8 & - & - \\
HiT-Base  ~\cite{kang2023hit} & ICCV2023 & - & 65.6 & 63.6 &  44.1    \\
HiT-Small  ~\cite{kang2023hit} & ICCV2023 & -  & 63.3 & 61.8 &  40.4  \\
HiT-Tiny  ~\cite{kang2023hit}  & ICCV2023 & - & 58.7 &   53.2 & 35.8   \\
MixformerV2-S  ~\cite{cui2024mixformerv2}  &NeurIPS2023 & 47.2 & 65.1 & 63.6 & -\\  

E.T.Track  ~\cite{ETTrack} &WACV2023 & -  & 62.3 & 59.0 & - \\

FEAR  ~\cite{borsuk2022fear} &ECCV2022 & - & 61.4 & - & - \\

HCAT  ~\cite{HCAT} & ECCVW2022 & - & 62.7 & 63.5 &  \\

LightTrack  ~\cite{lighttrack} &CVPR2021  & -&62.5 &55.3 &   -  \\



\bottomrule[1.5pt]
\end{tabular}
}
\label{tunl}
\end{table}

\subsection{Ablation Study} 

In this section, we present a comprehensive ablation study to analyze the effectiveness of our MST method. All models in the ablation experiments are trained and evaluated using the same dataset and hyperparameters to ensure a fair and consistent comparison.

\begin{table}
\caption{Ablation analysis of key components.}
\centering
\resizebox{1\linewidth}{!}{
\begin{tabular}{c|c|cc|c}
 \toprule[1.5pt]
\# & Method  &\begin{tabular}[c]{@{}c@{}} GOT-10K \\  \end{tabular} & \begin{tabular}[c]{@{}c@{}}  UAV123 \\  \end{tabular} &  $\triangle$ \\
 \midrule[0.75pt]

 \cellcolor{gray!20} \textbf{1} \cellcolor{gray!20} & \cellcolor{gray!20} \textbf{MST} \cellcolor{gray!20} & \cellcolor{gray!20}\textbf{69.6} \cellcolor{gray!20} & \cellcolor{gray!20}\textbf{68.4} \cellcolor{gray!20} &  \cellcolor{gray!20} \textbf{-} \cellcolor{gray!20} \\
 2 &  W/o SSE & 68.4 (-1.2) & 67.3 (-1.1) & -1.2 \\
 3 &  W/o CSI & 68.6 (-1.0) & 67.2 (-1.2) & -1.1 \\
 4 &  W/o SSE and CSI & 67.5 (-2.1) & 66.4 (-2.0)  & -2.1 \\
 5 &  W/o MSG, SSE and CSI & 67.6 (-2.0) & 66.2 (-2.2)  & -2.1 \\

\bottomrule[1.5pt]
\end{tabular}
}
\label{tab:as1}
\end{table}

\noindent\textbf{Analysis of Key Components.} To validate the effectiveness of each component in MST, we conducted a detailed quantitative ablation study, with the results presented in \Cref{tab:as1}. Configuration \#1 corresponds to the complete MST model, which integrates the MSG, SSE, and CSI modules to enhance feature representation. This full configuration achieves the best performance, significantly surpassing all other variants, demonstrating the overall strength of our design. When the SSE module is removed (Configuration \#2), while keeping all other settings identical, the performance drops by 1.2\% on the AO metric of GOT-10k and 1.1\% on the AUC metric of UAV123. Similarly, removing the CSI module (Configuration \#3) results in performance decreases of 1.0\% and 1.2\% on the respective datasets. These observations indicate that both SSE and CSI contribute substantially to the tracker's performance and exhibit a strong coupling effect that allows them to jointly improve MST’s tracking capability. Furthermore, Configuration \#4, which eliminates both SSE and CSI and simply aggregates multi-level features from the MSG module, shows a performance drop comparable to Configuration \#5, where all the proposed strategies are removed.The average score drop for both configurations is about 2.1\%, which further demonstrates the key role played by our proposed module in achieving robust and accurate tracking.

\begin{table}
\caption{The number of feature layers adopted by MST.}
\centering
\resizebox{1\linewidth}{!}{
\begin{tabular}{c|cc|c}
 \toprule[1.5pt]
 Number of layers  &\begin{tabular}[c]{@{}c@{}} GOT-10K \\  \end{tabular} & \begin{tabular}[c]{@{}c@{}}  UAV123 \\  \end{tabular} &  $\triangle$ \\
 \midrule[0.75pt]

 Baseline (w/o MSG, SSE, CSI) & 67.6 & 66.2  & - \\
 1 &  68.0 (+0.4) & 66.5 (+0.3) & +0.4 \\
 2 &  68.5 (+0.9) & 67.0 (+0.8) & +0.9 \\
 \cellcolor{gray!20} \textbf{3} \cellcolor{gray!20} &  \cellcolor{gray!20} \textbf{69.6 (+2.0)} \cellcolor{gray!20} & \cellcolor{gray!20} \textbf{68.4 (+2.2)} \cellcolor{gray!20} & \cellcolor{gray!20} \textbf{+2.1} \cellcolor{gray!20} \\
 4 &  69.0 (+1.4) & 67.6 (+1.4)  & +1.4 \\
 5 &  68.7 (+1.1) & 67.2 (+1.0)  & +1.1 \\

\bottomrule[1.5pt]
\end{tabular}
}
\label{tab:as2}
\end{table}

\noindent\textbf{The Number of Feature Layers Adopted by MST.} We conducted an ablation study to investigate the impact of the number of feature layers input into the SSE and CSI modules by MSG for specialized refinement and interaction enhancement. As shown in Table 5, when only one feature layer is used as input to the SSE and CSI modules, a modest performance improvement is observed, with the average increase of 0.4\% in the AO of GOT-10K and AUC of UAV123. We attribute this improvement to the further refinement of features by the SSE and CSI modules, which enhances their expressive capability to a small extent. When two feature layers are used, the performance improvement reaches 0.9\%, indicating that while two layers provide more diverse information compared to a single layer, the representation is still not fully optimized. With three feature layers, the average improvement increases to 2.1\%, suggesting that three layers offer a sufficient level of feature diversity and depth for effective enhancement. However, when four or five layers are used, the performance improvement plateaus, with increases of only 1.4\% and 1.1\%, respectively. This indicates that using more layers leads to confusion in the feature information, and the shallow features from the backbone network may introduce noise, interfering with the final feature representation.

\begin{table}
\caption{Analysis of modules composed of SSE and CSI.}
\centering
\resizebox{1\linewidth}{!}{
\begin{tabular}{c|cc|ccc}
\toprule[1.5pt]
Basic module &\begin{tabular}[c]{@{}c@{}} GOT-10k \\  \end{tabular} & \begin{tabular}[c]{@{}c@{}}  UAV123 \\  \end{tabular} &  FLOPs (G) & Param (M) & FPS \\
 \midrule[0.75pt]

Baseline &    67.6 & 66.2  & 2.18  & 7.14  &205/58 \\
 

 Bi-directional SSM &  68.7 (+1.1) & 67.7 (+1.5) & 2.34 (+0.16) & 7.57 (+0.40) &110/30 (-95/-23) \\
 HSM-SSD  & 68.5 (+0.9) & 67.6 (+1.4) &  2.26 (+0.08) & 7.74 (+0.60) &201/56 (-4/-2) \\
 \cellcolor{gray!20} \textbf{HSA-SSD} \cellcolor{gray!20}  & \cellcolor{gray!20} \textbf{69.6 (+2.0)} \cellcolor{gray!20} & \cellcolor{gray!20} \textbf{68.4 (+2.2)} \cellcolor{gray!20} & \cellcolor{gray!20}  \textbf{2.28 (+0.10)} \cellcolor{gray!20} & \cellcolor{gray!20} \textbf{7.80 (+0.66)} \cellcolor{gray!20} & \cellcolor{gray!20} \textbf{200/55 (-5/-3)} \cellcolor{gray!20} \\
 
\bottomrule[1.5pt]
\end{tabular}
}
\label{tab:as3}
\end{table}

\noindent\textbf{Analysis of SSE and CSI Variants.} To assess the impact of the SSE and CSI modules on tracking efficiency, we conducted an ablation study based on different implementations of their underlying models. Both SSE and CSI are designed to perform specialization and interactive aggregation of multi-level state features using lightweight state-space modeling. To this end, various types of state-space models can be adopted to realize these modules.

In \Cref{tab:as3}, the baseline refers to the version of our tracker without the SSE and CSI modules. We first implemented SSE and CSI using a Bi-directional State Space Model (Bi-SSM). Although this approach achieved the improvements of 1.1\% on the AO metric of GOT-10K and 1.5\% on the AUC of UAV123, it significantly degraded the inference speed on both GPU and CPU. Through code-level analysis, we attribute this slowdown to the inherently sequential nature of the bidirectional process, which hinders parallel computation. Moreover, we found that such bidirectional modeling is less suited for image sequences, contributing further to inefficiencies.

Next, we employed the HSM-SSD module from EfficientViM as a more efficient backbone for SSE and CSI. While the performance gains were slightly lower than those achieved with Bi-SSM, the inference speed remained nearly unaffected. Finally, we proposed an improved version of HSM-SSD, tailored specifically to handle diverse state features more effectively. The enhanced module delivered top performance, with the 2.0\% AO gain on GOT-10K and the 2.2\% AUC gain on UAV123,  while maintaining the same level of efficiency as the original HSM-SSD.

\noindent\textbf{Visualization Comparison.} To thoroughly assess the robustness of our MST tracker, we compare it against recent state-of-the-art lightweight trackers with similar computational complexity across a variety of challenging scenarios. In the first scene, where the tracked object is a bicycle undergoing significant background changes and interference from surrounding objects, MST consistently maintains accurate localization. In contrast, HiT-Small, HiT-Tiny, and HCAT exhibit varying degrees of drift or regression errors, failing to stably track the object.
The second and third scenarios involve severe distractor interference caused by similar-looking objects. In the second case, numerous similar blankets appear alongside drastic viewpoint shifts, while the third scene features multiple nearly identical goldfish in close proximity, making the tracking task highly ambiguous. Thanks to its ability to perform specialization and interactive aggregation of multi-level state features, MST is able to effectively distinguish the true target from the distractors, outperforming HiT-Small, HiT-Tiny, and HCAT in maintaining precise target tracking.
In the final scene, the target undergoes fast motion and is partially occluded by smoke, alongside the presence of similar distractors. MST remains the only tracker to accurately locate the object under occlusion, whereas HiT variants exhibit regression drift and HCAT misidentifies the target entirely. Furthermore, MST is the only tracker that continues to maintain stable and accurate tracking throughout the remainder of the sequence.

These qualitative comparisons across diverse real-world scenarios validate the effectiveness of the proposed SSE and CSI modules. Specifically, the SSE module specializes multi-state representations to better capture the unique characteristics of the target, while the CSI module enables effective information exchange across states. Together, they significantly enhance the representational capacity of the tracker, leading to more robust and reliable performance in complex environments.

\begin{figure}[t]
  \centering
   \includegraphics[width=1\linewidth]{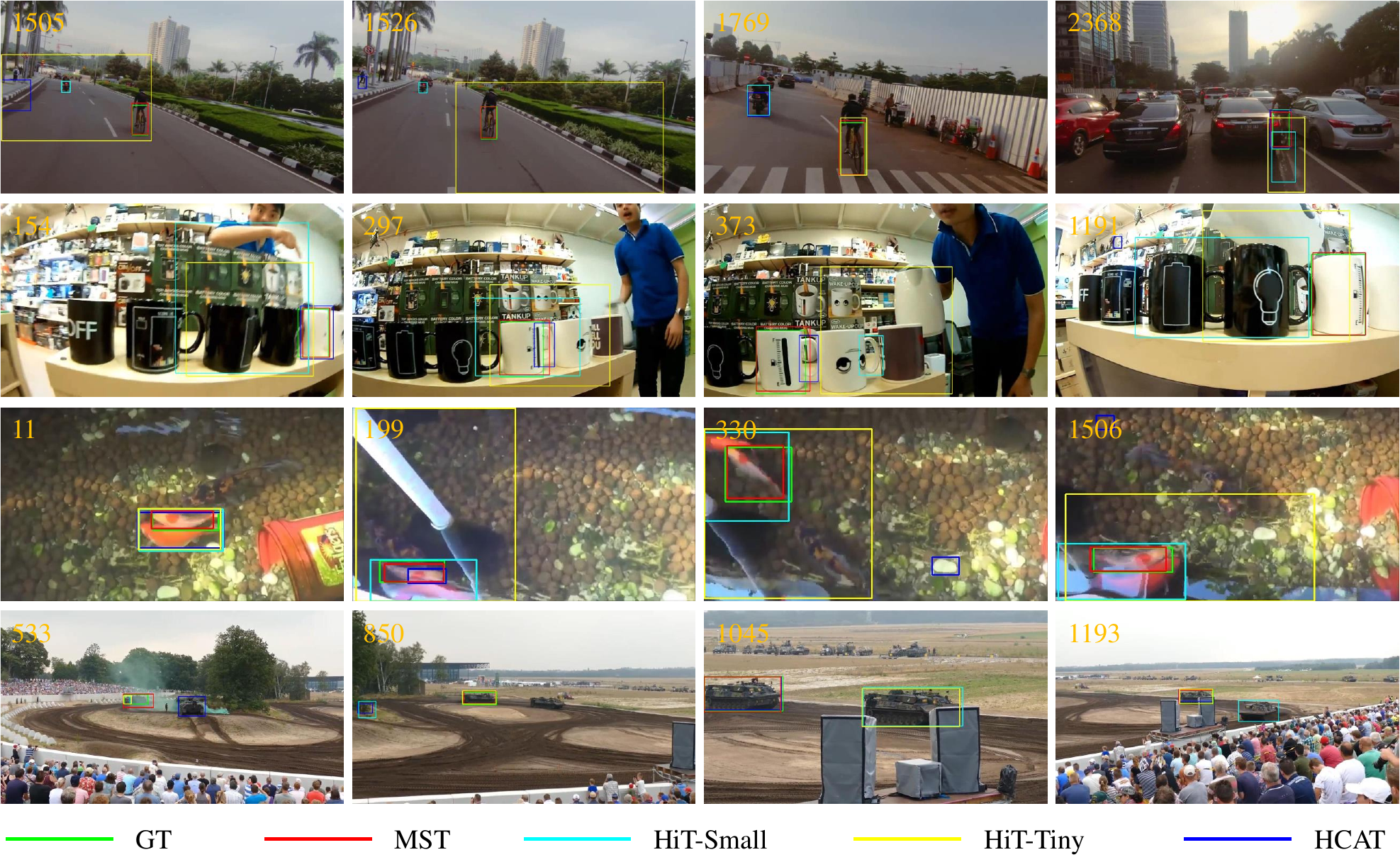}

   \caption{Visual comparison of tracking results.}
   \label{as:visual}
\end{figure}
\section{Conclusion}

In this paper, we introduced the Multi-State Tracker (MST), a novel and efficient tracking framework that enhances the robustness of lightweight trackers through specialized augmentation and interactive aggregation of multi-state representations. By integrating the proposed MSG, SSE, and CSI modules, MST effectively captures diverse and complementary characteristics of the target, enabling accurate tracking in complex scenarios such as occlusions, appearance variations, and motion blur. Notably, the SSE and CSI modules are designed with a lightweight HSA-SSD backbone, ensuring minimal computational overhead while delivering significant performance gains.
Extensive experiments on multiple challenging benchmarks, including GOT-10K, UAV123, and LaSOT, demonstrate that MST achieves state-of-the-art performance among efficient trackers, outperforming prior methods in both accuracy and speed. In particular, MST surpasses the leading efficient tracker HCAT by 4.5\% on GOT-10K while maintaining real-time performance on standard hardware.
Our results highlight the importance of multi-state representations in effective tracking and open new possibilities for deploying robust trackers in resource-constrained environments.

\begin{acks}
This work was supported in part by the National Natural Science Foundation of China under Grant 62376223, and in part by the fundamental Research Funds for the Central Universities.
\end{acks}

\bibliographystyle{ACM-Reference-Format}
\bibliography{ref}

\section*{Metrics Definitions}

In this section, we clearly define and explain each evaluation metric used in our experiments, both intuitively and mathematically. The metrics include Average Overlap (AO), Success Rate (SR), Precision (P), Normalized Precision (PNorm), and Area Under the Curve (AUC). 

\textbf{Average Overlap (AO):}

\[
\mathrm{AO} = \frac{1}{N} \sum _{i=1}^{N} \frac{|B_i^{\text{pred}} \cap B_i^{\text{gt}}|}{|B_i^{\text{pred}} \cup B_i^{\text{gt}}|}
\]

\textbf{Success Rate (SR$_\tau$) ($\tau = 0.5, 0.75$):}

\[
\mathrm{SR}_\tau = \frac{1}{N} \sum _{i=1}^{N} \mathbb{I} \left( \frac{|B_i^{\text{pred}} \cap B_i^{\text{gt}}|}{|B_i^{\text{pred}} \cup B_i^{\text{gt}}|} \geq \tau \right)
\]

\textbf{Precision (P$_\delta$) ($\delta = 20$):}

\[
\mathrm{P}_\delta = \frac{1}{N} \sum _{i=1}^{N} \mathbb{I} \left( \left\| c_i^{\text{pred}} - c_i^{\text{gt}} \right\|_2 \leq \delta \right)
\]

\textbf{Normalized Precision (PNorm$_\delta$):}

\[
\mathrm{PNorm}_\delta = \int _{0}^{0.5} \frac{1}{N} \sum _{i=1}^{N} \mathbb{I} \left( \frac{ \left\| c_i^{\text{pred}} - c_i^{\text{gt}} \right\|_2 }{ \sqrt{w_i^{\text{gt}} h_i^{\text{gt}}} } \leq d\delta \right)
\]

\textbf{Area Under the Curve (AUC):}

\[
\mathrm{AUC} = \int_{0}^{1} \mathrm{SR}_\tau \, d\tau
\]

\section*{Model Complexity Analysis}

We provide a detailed breakdown of FLOPs and parameters for each component to highlight the efficiency of our design, as shown in Table~\ref{tab:complexity}. From this breakdown, it is evident that the main computational cost lies in the lightweight backbone, while the SSE and CSI modules add only minimal overhead. This demonstrates that MST achieves richer target representations with negligible additional complexity, making it well suited for resource-constrained tracking scenarios.

\begin{table}[h]
\caption{FLOPs and parameter breakdown for each component in MST.}
\label{tab:complexity}
\centering
\begin{tabular}{lcc}
\toprule
Component & FLOPs (G) & Params (M) \\
\midrule
Backbone & 1.75 & 5.49 \\
SSE Module & 0.05 & 0.49 \\
CSI Module & 0.05 & 0.17 \\
Head & 0.53 & 2.31 \\
\bottomrule
\end{tabular}
\end{table}

\subsection*{Runtime Efficiency on Edge Devices}

We evaluate the runtime efficiency of MST on NVIDIA Jetson Xavier NX and compare it with other HiT variants. As shown in Table~\ref{tab:fps}, MST runs at 36 FPS, which is faster than HiT-Base (34 FPS), slightly slower than HiT-Small (37 FPS) and HiT-Tiny (41 FPS), while achieving the highest accuracy (69.6 AO on GOT-10k). This demonstrates that MST maintains strong real-time performance suitable for edge deployment scenarios, with a superior accuracy-efficiency trade-off compared to existing lightweight trackers.

\begin{table}[h]
\centering
\begin{tabular}{lcc}
\toprule
Method & FPS (Jetson Xavier NX) & AO (GOT-10k) \\
\midrule
HiT-Tiny & \textbf{41} & 52.6 \\
HiT-Small & 37 & 62.6 \\
HiT-Base & 34 & 64.0 \\
MST & 36 & \textbf{69.6} \\
\bottomrule
\end{tabular}
\caption{Comparison of runtime speed and accuracy (AO) on NVIDIA Jetson Xavier NX. MST achieves competitive FPS while offering the highest tracking accuracy.}
\label{tab:fps}
\end{table}

\end{document}